\documentclass{article}

\usepackage[includeheadfoot,
            bindingoffset=0mm,
            inner   = 25mm,
            top     = 10mm,
            outer   = 25mm,
            bottom  = 10mm,
            paperwidth = 210mm,
            paperheight = 297mm,
            ]{geometry}
\usepackage[margin={2.0cm,0cm},oneside,labelfont={sf,bf},singlelinecheck=false]{caption}

\usepackage{graphicx}

\usepackage[fleqn]{amsmath}
\usepackage{accents}
\usepackage{amssymb}

\usepackage{enumitem}
\usepackage{cite}
\usepackage[perpage,multiple]{footmisc}
\usepackage{hyphenat}
\usepackage[english]{babel}
\usepackage[section]{placeins}
\usepackage{color}
\usepackage{multirow}

\usepackage{titlesec}
\titleformat{\section}[hang]{\Large\bfseries\raggedright\sffamily}{\thesection}{1em}{}
\titleformat{\subsection}[hang]{\large\bfseries\raggedright\sffamily}{\thesubsection}{1em}{}
\titleformat{\subsubsection}[hang]{\normalsize\bfseries\raggedright\sffamily}{\thesubsubsection}{1em}{}
\usepackage{abstract}

\usepackage{authblk}

\begin{document}

\title{\huge\bfseries\sffamily Multivariate Quantile Regression via Kolmogorov-Arnold Networks}

\author[1]{Andrew Polar}
\author[2,3]{Michael Poluektov}
\affil[1]{Independent software consultant, Duluth, GA, USA}
\affil[2]{School of Computing and Mathematical Sciences, University of Greenwich, Park Row, London SE10 9LS, UK}
\affil[3]{Corresponding author, email: m.poluektov@greenwich.ac.uk}

\date{ \large\normalfont\sffamily DRAFT: \today }

\maketitle

\setlength{\absleftindent}{2.0cm}
\setlength{\absrightindent}{2.0cm}
\setlength{\absparindent}{0em}

\begin{abstract}
	This paper introduces a novel algorithm for predicting conditional joint distributions of vector-valued targets in stochastic systems whose randomness is intrinsic rather than arising from observation errors or additive noise. Multivariate quantile regression is based on modeling conditional joint distributions but represents a less challenging task. It predicts the probability that vector-valued targets fall within predefined regions, identifies regions corresponding to predefined probability levels, or performs both tasks simultaneously.
	
	The proposed identification technique employs ensembles of Kolmogorov--Arnold networks (KANs) as flexible function approximators. Although the suggested technique is not theoretically restricted to KANs, KANs are particularly well suited to the proposed construction and are therefore used throughout this study.
	
	In addition to the training procedure, this work introduces a new discrepancy measure for joint distributions and a goodness-of-fit (GoF) test based on it. This GoF test was initially developed to validate and calibrate the proposed identification technique and is used here in an ad hoc manner. Although the test could be tabulated for broader use, such a tabulation is not pursued in this work. The test is also applicable more generally.
	
\end{abstract}

\section{Introduction}
\label{sec:intro}

Methods of probabilistic modeling in the literature are divided into three groups \cite{gawlikowski2023survey}: Bayesian methods, ensemble methods, and test-time augmentation. This study introduces a new ensemble training approach for approximating distributions by a finite set of samples without any prior assumption about their type.

The list of such methods for scalar targets is not very long. It includes k-nearest neighbors (kNN), quantile regression forests (QRF) \cite{meinshausen2006quantile}, and, more recently, divisive data re-sorting (DDR) \cite{polar2025probabilistic}, published by the authors of this paper. Some other methods can be adapted to support conditional distributions of the scalar targets, including random forests \cite{Breiman2001}, XGBoost \cite{sluijterman2025composite}, CatBoost \cite{marz2020catboostlss}, kernel-based methods \cite{hyndman1996estimating}, and Gaussian processes \cite{dutordoir2018gaussian}.

Software packages implementing some of these methods are also available and can be used directly or readily customized for particular applications. 

One group of methods used for vector targets is based on dimensionality reduction \cite{galvao2025multivariate, kanazawa2026multivariate, Messoudi2021Copula}: copula-based methods, directional quantiles, and tomographic quantile forests. The basic concept is to replace vector targets with multiple scalar targets, for which univariate distributions are identified and then used to reconstruct quantiles. The reusable software is rarely provided \cite{kanazawa2026multivariate}, but still can be found.

The related topic in the literature is conformal prediction \cite{Klein2026Multivariate}, which uses calibration to construct prediction regions, whereas this study considers the prediction region to be specified by business requirements. 

In many cases, the multivariate quantle regressions use already established univatiate regression methods. The authors also follow this research policy. 
The suggested method is an extension of the previously published divisive data re-sorting (DDR) method \cite{polar2025probabilistic} to support vector targets and capture conditional joint probabilities. However, the research target is reduced to the simpler task of multivariate quantile regression.  

The success of the proposed modeling approach depends on one critical property: the conditional distribution should vary gradually with the observed features. Thus, substantially different feature vectors may correspond to substantially different conditional distributions, while small changes in the feature vector should result in correspondingly small changes in the distribution of the targets. Here, the terms \emph{small} and \emph{substantially} are used in an intuitive rather than strictly mathematical sense; a formal definition of this property is beyond the scope of the present study.

All experiments are fully reproducible, and the source code used in the study is publicly available.

\section{Projection Tree}
\label{sec:NK}

\begin{figure}[ht]
	\centering
	\begin{minipage}{0.49\textwidth}
		\centering
		\includegraphics[width=1.0\textwidth]{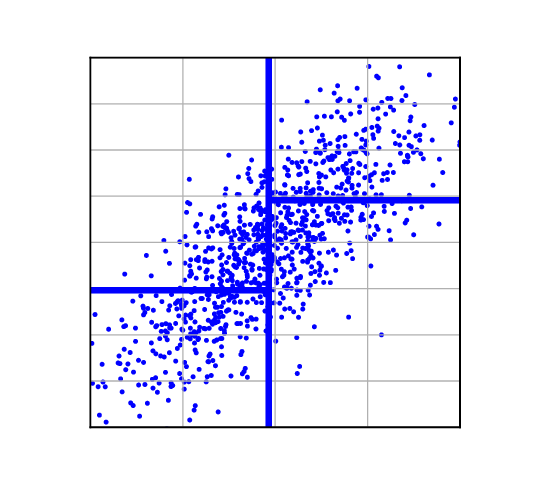}
	\end{minipage}
	\hfill
	\begin{minipage}{0.49\textwidth}
		\centering
		\includegraphics[width=1.0\textwidth]{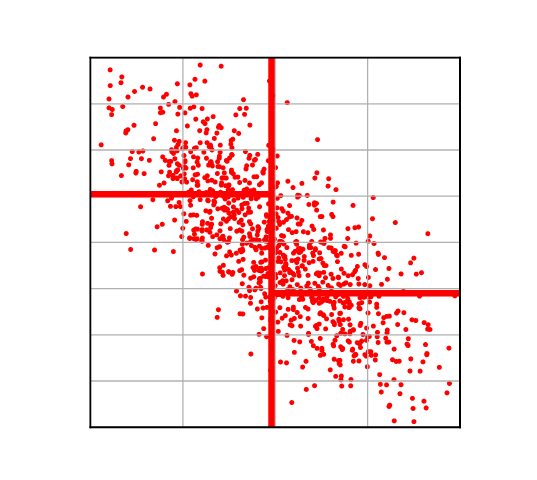}
	\end{minipage}
	\caption{Recursive median partitioning for two different joint distributions.}
	\label{fig:two_images}
\end{figure}

It is well established that marginal distributions do not provide an adequate representation of joint distributions in general. This is clearly illustrated in Figure~\ref{fig:two_images} by two samples: one red and one blue. Having an adequate discrepancy measure for joint distributions is critical for this research. The authors already introduced one measure --- recursive median partitioning or \textit{median tree} in the previous paper \cite{polar2025probabilistic} for univariate data and here it is generalized to multivariate case. 

A set of vectors $\{X_i\}$ in $\mathbb{R}^n$ is sorted according to the first component and then divided into two subsets at the median. Each of the resulting subsets is then sorted according to the second component and divided again at the corresponding median. This procedure continues sequentially through all components. After the last component has been used, the procedure returns to the first component and continues cyclically. The process terminates when the resulting subsets become sufficiently small or when targeted depth is achieved. The medians obtained during this recursive partitioning, recorded in the order in which they are generated, form a fingerprint of the multivariate distribution. 

The first two steps of such partitioning are illustrated in Figure~\ref{fig:two_images} by lines. It can be seen that median 
values for the first step (vertical lines) may be close, but other two, obtained for the left and right sides (horizontal lines), are significantly 
different. Continuation of partitioning will further exacerbate the divergence. For this particular case, depicted in Figure~\ref{fig:two_images}, only three numbers obtained by this rule are sufficient to reveal a significant difference in the joint distributions that is invisible in the marginal distributions.

Let $M_x$ and $M_y$ denote median vectors of the same depth for two corresponding sets of vectors. The distance between these median vectors can be used as measure. After adding a normalization it may be used as statistic $S$ in goodness-of-fit test: 
\begin{equation}
	S = 
	\frac{2\sqrt{2} \quad \lVert M_x - M_y \rVert}
	{\lVert M_x \rVert + \lVert M_y \rVert}
	\label{statistic}
\end{equation}
In the univariate case, the proposed statistic is related to the statistic used in the two-sample Cram\'er--von Mises test \cite{Anderson1962}. 

If ranges of the components vary significantly, the normalization should be applied prior to procedure. A different order in usage of components in the splitting process changes the resulting median vectors but does not change the principal. The distance between such vectors is a measure of discrepancy regardless of the order, provided that it is the same for both samples.

\subsection{Monte Carlo Simulation and $p$-value}
\label{sec:pvalue}

The median tree provides a statistical measure, but using it as a GoF test would ordinarily require its sampling distribution to be tabulated for different sample sizes and tree depths. For synthetic datasets, however, such tabulation is not necessary.

When the record with uncertainty is generated by computer code, it can be repeated to obtain a sufficiently large Monte Carlo population $P$ and a set of much smaller samples, for which the statistics $S_i$ can be found, using Equation \ref{statistic}, where one median vector is obtained for population $M_P$ and another for 
smaller sample $M_i$.

Same statistic for the predicted sample is matched with the sorted list of statistics obtained in the simulation for the distribution known to be the same and the conclusion based on a mechanism similar to a $p$-value can be made. 
This test can be considered an empirical/randomization $p$-value calibrated from Monte Carlo samples under the known reference population.

\subsection{Projection Vectors}
\label{sec:projection_trees}

The Orthogonal Projection Tree (OPT) introduced above is not the only possible choice. Some existing methods use randomly selected projection vectors, as in the Random Projection Tree (RPT) \cite{dasgupta2008random}. However, the goal of RPT is not to measure discrepancies between joint distributions, but rather to represent large sets of vectors more compactly by partitioning them into clusters and representing each cluster by its mean vector.

Several choices of projection vectors for different purposes are considered in this work. In addition to the OPT and RPT, the Principal Component Projection Tree (PCPT) is introduced, in which the eigenvector corresponding to the largest eigenvalue of the covariance matrix is used as the projection direction. This eigenvector defines the direction of maximum variance within the current subset of vectors. The covariance matrix and its principal eigenvector are recomputed for each subset as the partitioning proceeds recursively.

\section{The Magic of Calibration}

Monte Carlo sampling is not applicable to physical observations of real-world stochastic systems when repeated observations under exactly the same conditions are impossible. Nevertheless, the probabilistic accuracy of a model can still be assessed through calibration.

Suppose that, for each observation, the model generates a set of predicted target vectors and a region of interest. The fraction of predicted vectors that falls within the region provides an estimate of the probability.

When new experimental records become available, it can be recorded whether the actual target vector falls within the corresponding region. The fraction of such observations is the observed hit rate.

For a sufficiently large dataset, the observed hit rate should be close to the average of the estimated probabilities. In the ideal case, these two quantities are equal. This should hold even when different observations have different regions and different associated probabilities:

\begin{equation}
	\lim_{N\to\infty}
	\left(
	\frac{\sum_{i=1}^{N} P_i}{N}
	-
	\frac{N_{\mathrm{hits}}}{N}
	\right)
	= 0,
	\label{calibration}
\end{equation}
where $P_i$ is the probability estimated for the region associated with observation $i$, and $N_{\mathrm{hits}}$ is the number of observations for which the actual target vector falls within its corresponding region.

The principle is analogous to the law of large numbers at a roulette table in a casino. A guest may bet on a single cell, two cells, a row, four cells, a column, or larger sets of cells, including all even or all odd numbers. Each type of bet corresponds to a different probability of winning. Over a sufficiently large number of games, the ratio of the number of winning games to the total number of games should approach the corresponding average predicted probability.

Consequently, a sufficiently large calibration dataset that is not used for model training provides an opportunity to assess the probabilistic accuracy of the model. The calibration results can also be used for model selection and hyperparameter tuning. For example, if the probabilistic performance is unsatisfactory, the model architecture can be modified by adding or removing neurons or layers, after which the model can be retrained and evaluated again using the calibration dataset.

The calibration procedure therefore plays a role analogous to that of a validation set in deterministic machine learning. When the end user of the model is interested only in a specific region or quantile, the model can be tuned to perform well for this particular choices while ignoring other regions or quantiles. This creates an opportunity to use smaller, narrowly customized models.

\section{Multivariate Divisive Data Re-sorting (MDDR)}
\label{MDDR}

\subsection{Kolmogorov--Arnold Networks (KANs)}
\label{NK}

The proposed technique trains an ensemble of deterministic models. In principle, the individual models are not restricted to a particular model class. However, all experiments presented in this study use Kolmogorov--Arnold networks (KANs), and the method is therefore described and evaluated in this concrete implementation. The reason of this choice is very banal: the authors have used KANs since 2021 \cite{polar2020deep} and are comfortable with its implementation. There is also a technical reason: recent research by the authors \cite{polar2026concurrent} has shown that KANs can be trained concurrently and can achieve substantial speed improvements compared with software implementations based on multilayer perceptrons (MLPs). Since the number of models in ensemble may exceed hundred, computational speed is therefore important. 

The core element of a KAN is a small generalized additive model (GAM), which plays a role analogous to that of a neuron in a conventional neural network:
\begin{equation}
	z = \alpha \sum_{j=1}^{m} g_{j}\left(x_j\right) + \beta,
	\label{urysohn}
\end{equation}
where $x_j$ are the inputs, $z$ is the output, $g_j$ are trainable functions, and $\alpha$ and $\beta$ are scaling parameters. Prior to the work presented in \cite{Poluektov2020}, training GAMs of this type was computationally expensive. The approach proposed in \cite{Poluektov2020} provides an efficient alternative and allows this element to be used as a node in the large tree, which is KAN. During training, backpropagation determines an improving increment $\Delta z$, which is used to update the functions $g_j$ so that individual GAM produces an improved estimate $z+\Delta z$. Parameters $\alpha$ and $\beta$ are 1 and 0 at the start and also adjusted in training. The only reason for $\alpha$ and $\beta$ is maintaining 
output $z$ in the predefined domain, since it is an argument of another function in a tree. 

\subsection{Training the Ensemble: the Core of the Method}

The training procedure has a conceptual similarity to the median tree described in Section~\ref{sec:NK}. First, a deterministic model is trained using the entire dataset. The vectors of residual errors are then calculated, projected onto vector, sorted, and used to divide the records into two subsets at the median. The same procedure is subsequently applied to each resulting subset. When binary vectors are used, the position of the $1$ is shifted cyclically at each step.

Depending on the size of a subset, a new model can be initialized for that subset, or an existing model can be copied and further trained on the corresponding data. An important practical property of KANs, reported by multiple researchers \cite{Deventer2022}, is their ability to retain previously learned behavior while adapting to new data. This property is, of course, mentioned in comparison with conventional neural networks. \textit{Forgetting} previously learned behavior is part of any iterative method to a certain degree. When the divided blocks become small, retaining previously learned behavior becomes important. At each previous step, the model is trained on both parts, so moving to a partial list of records may be regarded as narrowing the \textit{learning skills} available to the model.

After each split, the parent models and their corresponding datasets are no longer required. Only the models and data associated with the current leaf nodes need to be retained. Consequently, the procedure does not explicitly construct or store a complete tree. Instead, the number of active leaf models doubles at each splitting stage, and these models can be trained concurrently.

The final ensemble consists of the models associated with the leaf nodes. The training records themselves are neither copied nor physically moved between nodes; each leaf requires only a reference to the corresponding subset of records. Since the leaf models are trained on disjoint subsets, their training can be performed in parallel. This approach can be used to optimize the training of large models on large datasets.

\begin{figure}
	\begin{center}
		\includegraphics[width=0.95\textwidth]{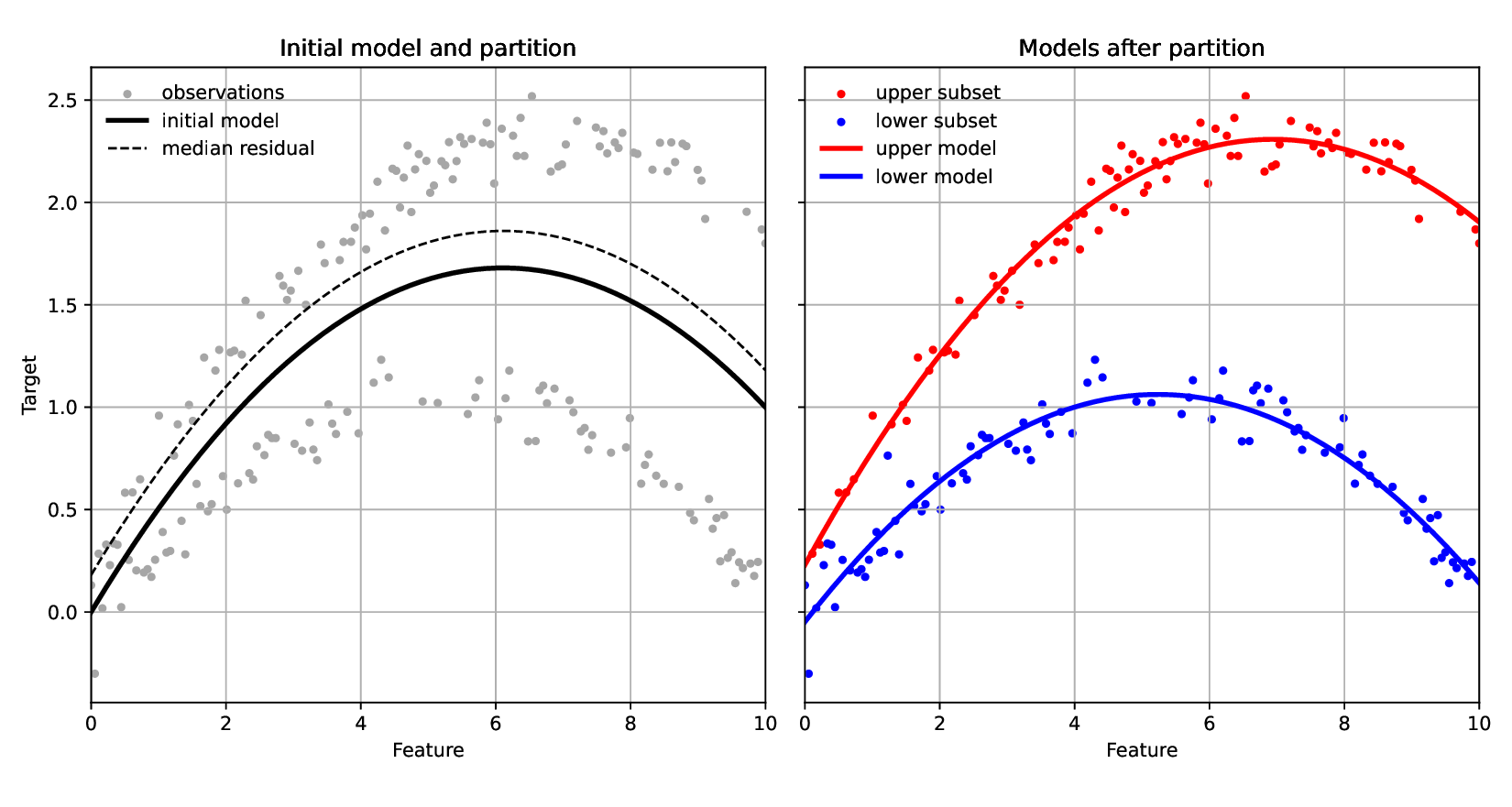}
	\end{center}
	\caption{One step of DDR logic illustrated by 1D example.}
	\label{fig:parabola}
\end{figure}

Figure~\ref{fig:parabola} illustrates how these steps can lead to a model of a joint distribution using a simple example. The observation points for a 1D model are shown in the left image. The initial model is chosen as a parabola and obtained by minimizing the residuals, shown as a solid black line. The dotted line divides the records according to their residual errors. An ensemble of two parabolas is then obtained for the two resulting subsets and shown in different colors in the right image. Some data properties are exaggerated for illustrative purposes to make the concept clear. For this reason, the illustration is limited to a 1D case with a visible pattern in the data. Even with a relatively small predicted sample, the two models capture the principal pattern of the uncertainty.

The choice of projection vectors may depend on the specific characteristics of the data and the modeling objectives. Three previously mentioned approaches are investigated in this study: orthogonal, random, and principal component-based projections.

\section{Experimental Part}
\label{sec:experimental}

\subsection{Product of Complex Numbers: Testing the Distribution}
\label{complex}

\begin{figure}
	\begin{center}
		\includegraphics[width=0.4\textwidth]{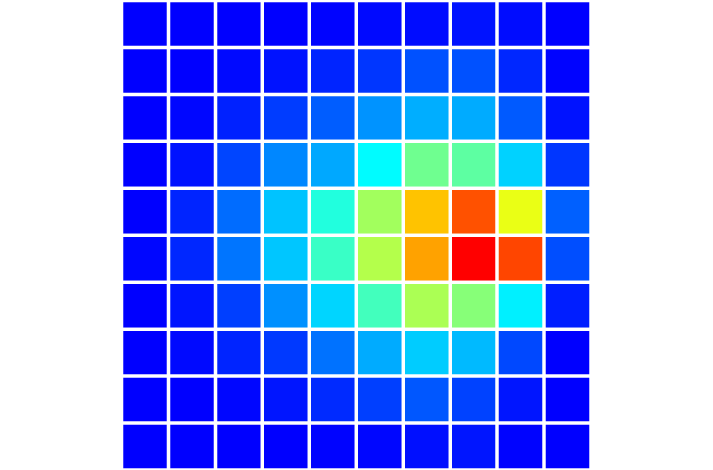}
	\end{center}
	\caption{The tiles presentation of the histogram for joint distribution of the product of two complex numbers with uncertainty. Each tile 
	represent frequency for the bucket, blue color is used for lower values and red color for higher.}
	\label{fig:tiles_complex}
\end{figure}

The first synthetic dataset is built as the product of two complex numbers, with four features and two jointly distributed targets. Unobserved uniformly distributed errors with zero mean were added to the real and imaginary parts of each factor before computing the products. The range of the errors is set to 50\% of the range of the observed components. The exact part of the features was also generated using a uniform distribution. In the Monte Carlo simulation, the observed part was kept constant while the errors were refreshed at each step to generate a new target. One example of such a target distribution is shown in Figure~\ref{fig:tiles_complex}. The frequencies in buckets of constructed histogram are expressed by tiles' colors, the lower in blue, the higher in red.

The baseline test was kNN. It was compared with MDDR as described in Section~\ref{sec:NK}. Two dataset sizes were used: 10,000 and 20,000 records. The target sample size was 128 for both kNN and MDDR. The GoF test was based on median trees, with the samples divided at the median until at least 3 records remained. The test was considered passed at the 1\% significance level using ad hoc generated $p$-values, as described in Subsection~\ref{sec:pvalue}. The data were generated anew at each execution for both kNN and MDDR. This explains the correlation between the numbers of passed tests.

The test results are shown in Tables~\ref{tab:10,000} and ~\ref{tab:20,000}. For both kNN and MDDR, the number of passed tests depends on the dataset size, as expected. The accuracy of MDDR is significantly higher in all tests. It can also be noted that the data-generation procedure is favorable to kNN. The observed features are drawn from a uniform distribution, and the unobserved errors are also uniformly distributed. Consequently, the used dataset has approximately uniform density without gaps, which is favorable for kNN.

The KAN model has layers and is defined by the sizes of the mapped vectors. In this case, the architecture was $[4,8,2]$, where the first layer maps a vector of 4 features into 8 hidden values, and the second layer maps these 8 values to the targets. The complexity of each function is defined by the number of points. For this model, it was $[4,8]$. All functions in the first layer were defined by 4 values, and all functions in the second layer by 8 values at evenly spaced argument values. All functions were piecewise linear. The total number of parameters for this model is 256.

The data records are divided until the remaining size is still greater than 50, which makes the last size close to 78. The model that is retrained on 78 records is a copy of the model trained in the previous leaf, which has double this number of records. This pattern exists throughout the entire training process, starting from either 10,000 or 20,000 records for the same model size.

In addition to the proposed goodness-of-fit (GoF) test, a widely used statistical measure, Energy Distance, was also employed \cite{szekely2013energy}. Energy Distance was computed between each Monte Carlo population and its corresponding model sample. The mean values across 100 kNN tests were $0.0490$ and $0.0276$ for datasets containing 10K and 20K records, respectively. For MDDR, the corresponding mean values were $0.0166$ and $0.0128$, respectively.

The advantage of MDDR over kNN is not limited to accuracy. The model also compensates for gaps in data density and eliminates the need to retain a large dataset for prediction.

Source code is available\footnote{https://github.com/andrewpolar/Cloud}.

\begin{table}
	\begin{center}
		\caption{Goodness-of-fit tests for joint distribution, dataset is 10,000 records.}
		\label{tab:10,000}
		\begin{tabular}{| l | c c c c c c c c |}
			\hline
			Code executions & $1$ & $2$ & $3$ & $4$ & $5$ & $6$ & $7$ & $8$ \\
			\hline
			GoF tests passed for kNN from 100 & $32$ & $26$ & $22$ & $24$ & $29$ & $21$ & $28$ & $24$ \\
			\hline
			GoF tests passed for MDDR from 100 & $49$ & $42$ & $37$ & $51$ & $55$ & $41$ & $39$ & $50$ \\
			\hline
		\end{tabular}
	\end{center}
\end{table}

\begin{table}
	\begin{center}
		\caption{Goodness-of-fit tests for joint distribution, dataset is 20,000 records.}
		\label{tab:20,000}
		\begin{tabular}{| l | c c c c c c c c |}
			\hline
			Code executions & $1$ & $2$ & $3$ & $4$ & $5$ & $6$ & $7$ & $8$ \\
			\hline
			GoF tests passed for kNN from 100 & $37$ & $38$ & $51$ & $31$ & $43$ & $44$ & $46$ & $44$ \\
			\hline
			GoF tests passed for MDDR from 100 & $62$ & $51$ & $70$ & $43$ & $72$ & $73$ & $56$ & $50$ \\
			\hline
		\end{tabular}
	\end{center}
\end{table}

\subsection{Dice Sets: Testing the Calibration}
\label{dice}
\begin{figure}
	\begin{center}
		\includegraphics[width=0.27\textwidth]{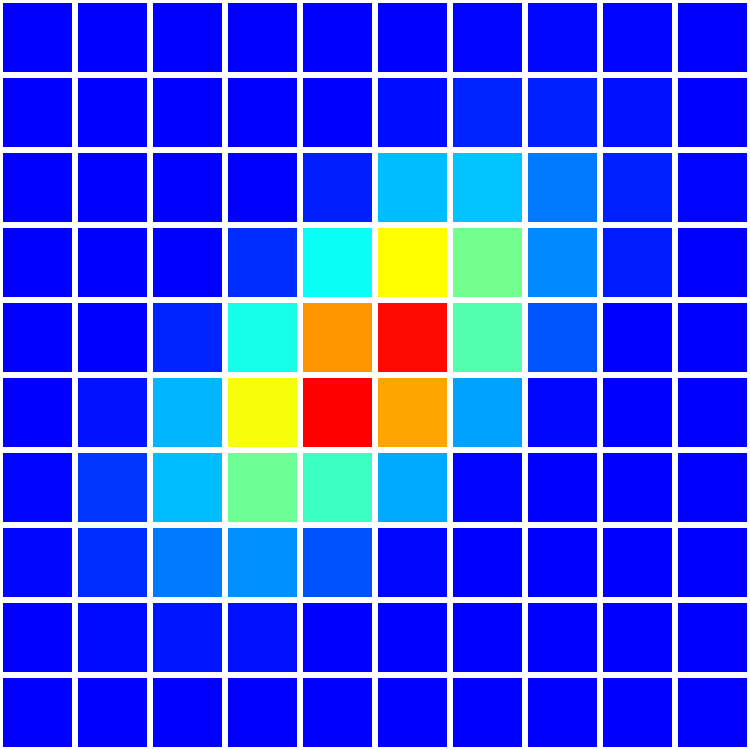}
	\end{center}
	\caption{The tiles presentation of the histogram for joint distribution of sums and differences of two dice quantities. Each tile 
		represents frequency for the bucket, blue color is used for lower values and red color for higher.}
	\label{fig:tiles_dice}
\end{figure} 

The features in this test are the numbers of dice $q_1, q_2, \ldots, q_n$. Each roll of these sets generates the same number of hidden variables $h_1, h_2, \ldots, h_n$, defined as the sums of all outcomes within the corresponding groups. These random variables are mapped to target vectors in two different ways: by multiplication with a matrix $L \in \mathbb{R}^{m \times n}$ or by computing multiple quadratic forms with symmetric matrices $Q_1, Q_2, \ldots, Q_m$. In this way, the numbers of features and targets are assigned in the code and can be varied to increase the data complexity and the difficulty of the training task.

The distributions of the hidden variables are non-Gaussian. For $q=1$, the distribution is uniform; for $q=2$, it is triangular; for larger values of $q$, it approaches a normal distribution according to the central limit theorem. Consequently, their different linear combinations may be regarded as non-Gaussian distributions that are nevertheless close to mixture of Gaussians. Nonlinear mapping via quadratic forms makes the resulting joint distribution substantially less Gaussian and makes the training task much more challenging. One very basic example is shown in Figure~\ref{fig:tiles_dice} as colored tiles depicting the frequencies of the buckets. It has only two targets, computed as the sum $h_1+h_2$ and the difference $h_1-h_2$ of two hidden variables.

The dice are virtual, with outcomes in the range $[1,10]$. The number of dice in each group also ranges from $1$ to $10$, so each hidden variable takes values in the range $[1,100]$.

The aim of this experiment is to evaluate the calibration accuracy of the proposed method. The code is publicly available\footnote{https://bitbucket.org/kolmogorov-arnold/dice3/src/main/} and can be used to generate a variety of datasets. Therefore, only one example is presented in detail in this paper:
\begin{center}
	\begin{tabular}{ll}
		\textbf{Training set}: & 10,000 \\
		\textbf{Calibration set}: & 10,000 \\
		\textbf{Number of features}: & 8 \\
		\textbf{Number of targets}: & 4 \\
		\textbf{Type of target mapping}: & quadratic forms \\
		\textbf{Size of each model}: & 300 trainable parameters \\
		\textbf{Size of ensemble}: & 256.
	\end{tabular}
\end{center}
The Pearson correlations among the four targets are shown in the matrix below.
\[
\begin{array}{cccccccc}
	& 0.54  & 0.14  & 0.12 \\
	&       & -0.14 & -0.32 \\
	&       &       & 0.68
\end{array}
\]

The region of interest is chosen as a hypersphere centered at the median point. This median point is determined independently for each target component and therefore does not necessarily correspond to an observed sample point. The radius of the hypersphere is selected to enclose a specified proportion of the generated target samples.

Three different projection choices OPT, RPT and PCPT, mentioned in Subsection \ref{sec:projection_trees} were tested. The results are shown in Table~\ref{tab:calibration}
along with baseline kNN test. The nonlinear mapping of the hidden outcomes of dice rolls to four targets represents a substantially more challenging case for regional calibration.
On that reason kNN performs less accurately than for the products of two complex numbers with unobserved disturbances in their factors.

\begin{table}[ht]
	\centering
	\caption{Calibration results for MDDR and kNN.}
	\label{tab:calibration}
	\renewcommand{\arraystretch}{1.4}
	\begin{tabular}{c|ccc|c}
		\hline
		\textbf{Target probability} &
		\multicolumn{3}{c|}{\textbf{MDDR}} &
		\textbf{kNN} \\
		\cline{2-4}
		&
		\textbf{OPT} &
		\textbf{RPT} &
		\textbf{PCPT} &
		\\
		\hline
		0.10 & 0.13 & 0.11 & 0.12 & 0.28 \\
		0.20 & 0.24 & 0.22 & 0.22 & 0.46 \\
		0.30 & 0.34 & 0.32 & 0.33 & 0.58 \\
		0.40 & 0.43 & 0.41 & 0.42 & 0.68 \\
		0.50 & 0.52 & 0.50 & 0.51 & 0.75 \\
		0.60 & 0.60 & 0.59 & 0.61 & 0.83 \\
		0.70 & 0.69 & 0.68 & 0.70 & 0.89 \\
		0.80 & 0.77 & 0.76 & 0.78 & 0.93 \\
		0.90 & 0.87 & 0.85 & 0.88 & 0.97 \\
		\hline
	\end{tabular}
\end{table}

The dataset is divided eight times during training, resulting in final subsets containing approximately 39 records each. The model has 300 trainable parameters. However, training starts with the entire dataset of 10,000 records, after which two copies of the model continue training on disjoint subsets of 5,000 records, and so on. The training method is Newton--Kaczmarz \cite{Poluektov2023}, which represents iterative improvements by compensating for backpropagation errors, as briefly explained in Subsection~\ref{NK}. The number of epochs after each split controls how the model is adapted from its previously learned properties. This parameter insignificantly contributes to the accuracy of the trained ensemble.

There is a principal difference between retraining a model on new data and the MDDR procedure. When a model is trained on a large dataset and then continues training on new records, previously learned properties may be lost. But in this test all models are trained on the same records multiple times, starting from the very first step. This explains why subsequent training of models on smaller subsets does not lead to overfitting. This logic can be summarized simply as skipping more and more records in subsequent epochs, starting with the entire dataset and gradually transitioning to multiple models trained on specifically customized subsets. 

Adjusting hyperparameters using a calibration data, including number of epochs in subsets, makes the calibration records part of the training procedure, and evaluating accuracy therefore requires new data. This is a common practice in which three datasets are used instead of two. In this experiment, however, a new dataset is generated for each execution, so a separate third dataset is not required.

Based on this test, it is difficult to determine the best projection type. The errors reported in Table~\ref{tab:calibration} are small, and slight changes in the hyperparameters may improve one category while worsening another. Based on the authors' experience, the random projection tree is recommended as the most convenient option to implement.

\subsection{Real-World Application}
\label{sec:real}
The dataset is taken from the UCI Archive\footnote{https://archive.ics.uci.edu/dataset/235/individual+household+electric+power+consumption} and represents a time series of electric power consumption in a single household, recorded at one-minute intervals over a period of almost four years. A few sample records are shown below:

\begin{verbatim}
16/12/2006;18:33:00;2.714;0.162;234.820;11.600;0.000;0.000;17.000
16/12/2006;18:34:00;3.538;0.086;233.760;15.600;0.000;1.000;16.000
16/12/2006;18:35:00;6.072;0.000;232.480;26.400;0.000;27.000;17.000
16/12/2006;18:36:00;4.536;0.000;233.540;19.400;0.000;1.000;17.000
\end{verbatim}

Each subsequent record is correlated with the previous one, and deterministic models for predicting the next record achieve nearly 90\% accuracy. However, since predicting power consumption for the next minute may have limited practical value, the data are aggregated into 15-minute blocks. The entire dataset contains 136,618 such blocks. Half of the blocks are used for training, the next quarter for calibration, and the remaining quarter for testing. The number of features is 10, including month, day of the week, and hour as additional features. The number of targets is 7, their correlation matrix is shown below. 
\[
\begin{array}{cccccccc}
	&        & 0.25  & -0.40 & 1.00  & 0.48  & 0.43  & 0.64 \\
	&        &       & -0.11 & 0.27  & 0.12  & 0.14  & 0.09 \\
	&        &       &       & -0.41 & -0.20 & -0.17 & -0.27 \\
	&        &       &       &       & 0.49  & 0.44  & 0.63 \\
	&        &       &       &       &       & 0.05  & 0.10 \\
	&        &       &       &       &       &       & 0.08
\end{array}
\]

The results from four executions of the code are presented in Table~\ref{tab:experimental}.

\begin{table}[ht]
	\centering
	\caption{Testing results for electric power consumption.}
	\label{tab:experimental}
	\begin{tabular}{c|cccc}
		\hline
		\textbf{Target probability} & \textbf{Actual 1} & \textbf{Actual 2} & \textbf{Actual 3} & \textbf{Actual 4} \\
		\hline
		0.10 & 0.11 & 0.11 & 0.11 & 0.10 \\
		0.20 & 0.23 & 0.22 & 0.18 & 0.19 \\
		0.30 & 0.33 & 0.33 & 0.27 & 0.28 \\
		0.40 & 0.43 & 0.45 & 0.36 & 0.37 \\
		0.50 & 0.53 & 0.56 & 0.45 & 0.47 \\
		0.60 & 0.63 & 0.67 & 0.56 & 0.59 \\
		0.70 & 0.74 & 0.77 & 0.68 & 0.70 \\
		0.80 & 0.84 & 0.87 & 0.81 & 0.83 \\
		0.90 & 0.92 & 0.94 & 0.92 & 0.93 \\
		\hline
	\end{tabular}
\end{table}

The feature reduction technique is used in the code primarily for demonstration purposes, since it was not strictly necessary for this dataset. The KAN model provides a vector-to-vector mapping similar to a neural network and can be trained as a two-layer deterministic model. The inner layer can then be used for feature reduction, while the outer layer is discarded. The output of the inner layer is subsequently used as the input for MDDR.

Several experiments conducted by the authors on different datasets have shown that, when applied correctly, this approach does not lead to a degradation in accuracy. It may also facilitate the construction of ensembles for datasets with long feature vectors and relatively small numbers of records.

Since household power consumption is related to the social activity of its inhabitants while remaining correlated with recent activity and date/time parameters, and given the size of the target vector, the authors consider these results satisfactory.

The code, along with the dataset provided for convenience, is publicly available\footnote{https://github.com/andrewpolar/MDDR}.

\section{Conclusions}
\label{sec:conclusion}

This article introduces and validates several concepts that are simple and intuitive. One of these is the median-tree statistic, in which a sequence of median values provides a compact fingerprint of a joint probability distribution. Another is a related method of reorganizing records during the training of an ensemble of models.

The proposed MDDR method provides practical advantages in addition to its accuracy. Even if MDDR produced models with the same accuracy as kNN, holding and using a compact ensemble of trained models would be more convenient than using entire large dataset for prediction. In the experiments presented here, however, MDDR also demonstrated a substantial accuracy advantage over kNN.

The implementation producing the presented results is short, does not depend on third-party libraries, and can be understood and customized by engineers for specific applications. 

Calibration of probabilistic models is a well-established practice. This work demonstrates that accurate calibration can be achieved even for stochastic systems with substantial uncertainty and related target components. 

\bibliographystyle{unsrt}
\bibliography{refs}

\end{document}